\documentclass{article}

\usepackage{arxiv}

\usepackage[utf8]{inputenc} 
\usepackage[T1]{fontenc}    
\usepackage{hyperref}       
\usepackage{url}            
\usepackage{booktabs}       
\usepackage{amsfonts}       
\usepackage{nicefrac}       
\usepackage{microtype}      
\usepackage{graphicx}
\usepackage{natbib}
\usepackage{doi}

\usepackage{epsfig}
\usepackage{subcaption}
\usepackage{calc}
\usepackage{amssymb}
\usepackage{amstext}
\usepackage{amsmath}
\usepackage{amsthm}
\usepackage{multicol}
\usepackage{pslatex}
\usepackage{algorithm2e}
\usepackage[bottom]{footmisc}
\usepackage{adjustbox}
\usepackage{pgf}

\title{Semantic Textual Similarity Assessment in Chest X-ray Reports Using a Domain-Specific Cosine-Based Metric}

\date{February, 2024} 					
\author{ \href{https://orcid.org/0000-0003-2675-5463}{\includegraphics[scale=0.06]{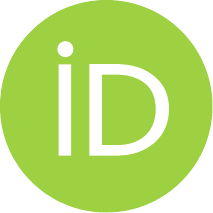}\hspace{1mm}Sayeh GHOLIPOUR PICHA}\\
	Univ. Grenoble Alpes,\\
        CNRS, Grenoble INP, GIPSA-lab,\\
        38000 Grenoble, France\\
	\texttt{sayeh.gholipour-picha@grenoble-inp.fr} \\
	\And
	\href{https://orcid.org/0000-0002-6258-6970}{\includegraphics[scale=0.06]{orcid.pdf}\hspace{1mm}Dawood AL CHANTI} \\
	Univ. Grenoble Alpes,\\
        CNRS, Grenoble INP, GIPSA-lab,\\
        38000 Grenoble, France\\
	\texttt{dawood.al-chanti@grenoble-inp.fr} \\
        \And
        \href{https://orcid.org/0000-0002-5937-4627}{\includegraphics[scale=0.06]{orcid.pdf}\hspace{1mm}Alice CAPLIER} \\
	Univ. Grenoble Alpes,\\
        CNRS, Grenoble INP, GIPSA-lab,\\
        38000 Grenoble, France\\
	\texttt{alice.caplier@grenoble-inp.fr} \\
}

\renewcommand{\headeright}{}

\hypersetup{
pdftitle={Semantic Textual Similarity Assessment in Chest X-ray Reports Using a Domain-Specific Cosine-Based Metric.},
pdfsubject={cs.CV},
pdfauthor={Sayeh GHOLIPOUR PICHA, Dawood AL CHANTI, Alice CAPLIER},
pdfkeywords={Semantic Similarity, Medical Language Processing, Biomedical Metric.},
}
\begin{document}
\maketitle
\begin{abstract}
Medical language processing and deep learning techniques have emerged as critical tools for improving healthcare, particularly in the analysis of medical imaging and medical text data. These multimodal data fusion techniques help to improve the interpretation of medical imaging and lead to increased diagnostic accuracy, informed clinical decisions, and improved patient outcomes. The success of these models relies on the ability to extract and consolidate semantic information from clinical text. This paper addresses the need for more robust methods to evaluate the semantic content of medical reports. Conventional natural language processing approaches and metrics are initially designed for considering the semantic context in the natural language domain and machine translation, often failing to capture the complex semantic meanings inherent in medical content. In this study, we introduce a novel approach designed specifically for assessing the semantic similarity between generated medical reports and the ground truth. Our approach is validated, demonstrating its efficiency in assessing domain-specific semantic similarity within medical contexts. By applying our metric to state-of-the-art Chest X-ray report generation models, we obtain results that not only align with conventional metrics but also provide more contextually meaningful scores in the considered medical domain.
\end{abstract}


\section{\uppercase{Introduction}}
\label{sec:introduction}

Advancements in deep learning for medical language processing have significantly improved healthcare clinical analysis, particularly in the domain of medical imaging applications. Notably, there has been substantial progress in generating chest X-ray reports comparable to those written by radiologists. However, a critical challenge persists in the chest X-ray application—assessing the semantic similarity between generated reports and the ground truth.

Identifying semantic similarities in medical texts is a difficult task within the language processing domain \cite{intro1}. 
This task necessitates a comprehensive grasp of the entire medical text corpus, the ability to recognize key content, and a profound understanding of the semantic relationships between these critical keywords at an expert level.
While existing metrics and approaches for capturing semantic similarity in natural language are effective, they are not designed for the complexities of medical content. The need for a robust metric to assess semantic similarity in medical texts has become increasingly evident, particularly in applications like chest X-ray report generation, and continues to be an active area of research \cite{cxr-repair-endo21a}, \cite{miura-etal-2021-improving-m2trans}, \cite{yuEvaluatingProgressAutomatic2022}.

State-of-the-art chest X-ray report generation models \cite{chen2022generating}, \cite{miura-etal-2021-improving-m2trans}, \cite{cxr-repair-endo21a} still rely on conventional Natural Language Processing (NLP) methods like BLEU \cite{papineni-etal-2002-bleu}, METEOR \cite{banerjee-lavie-2005-meteor}, and ROUGE \cite{lin-2004-rouge} to evaluate the generated reports against ground truth references. However, these metrics produce unreliable results due to their inability to comprehend and compare the semantic similarity of key medical terms.
A medical semantic similarity metric would not only provide more significant evaluation scores but could also be incorporated into the training process to improve model performance, potentially leading to enhanced diagnostic accuracy and decision-making. Additionally, as part of our ongoing research, our goal is to focus on providing visual interpretations of chest X-ray reports using text-to-image localization. As a consequence, a robust semantic similarity evaluation metric suitable for medical content will ensure the reliability of generated reports and will enable us to achieve more accurate localization and interpretation of image content.

In this context, we propose a new metric designed to assess and assign scores about the semantic similarity of medical texts. Our metric consists of two sequential steps: first, we identify the primary clinical entities, and subsequently, we evaluate the similarity between these entities using the domain-specific Cosine similarity score. Notably, our approach considers the presence of negations and detailed descriptions associated with medical entities during the evaluation process. To this end, our contributions include:

\begin{itemize}
    \item Introduction of a novel system for clinical entity extraction from medical texts.
    \item Proposition of a new scoring system for the evaluation of semantic similarity that suits medical and natural texts.
    \item Presentation of a validation method for scoring verification.
\end{itemize}

This paper is structured as follows: Section \ref{sec:related} discusses related works; Section \ref{sec:method} presents the theoretical and mathematical part of the novel metric; Section \ref{sec:validation} validates the metric; Section \ref{sec:result} discusses the results; Finally, Section \ref{sec:conclusion} concludes the paper.

\section{\uppercase{Related Works}}
\label{sec:related}

Recent studies have addressed the challenge of similarity evaluation between generated medical reports and the ground truth through various approaches other than conventional NLP metrics. Researchers have often introduced innovative metrics in the process.

In the CXR-RePaiR model by Endo et al. \cite{cxr-repair-endo21a} a unique approach for automatically evaluating chest X-ray report generation is proposed by introducing the CheXbert vector similarity metric, using the CheXbert labeler \cite{Smit2020CheXbertCA} — a specialized tool for chest X-ray report labeling. The process involves extracting labels from generated reports, comparing them with ground truth labels, and presenting the final score using cosine similarity. While this approach outperforms the BLEU metric, its applicability is limited to the specific context of chest X-ray reports and does not readily extend to other medical applications. 
The limitations arise from Chexbert being exclusively trained for chest X-ray reports. Moreover, the Chexpert labels \cite{irvin2019chexpert} (Atelectasis, Cardiomegaly, Consolidation, Edema, Enlarged Cardiomediastinum, Fracture, Lung Lesion, Lung Opacity, No Finding, Pleural Effusion, Pleural Other, Pneumonia, Pneumothorax) are specific to the chest X-ray dataset, further limiting the generalizability of the approach to other medical contexts.

In a separate study, Yu et al. \cite{yuEvaluatingProgressAutomatic2022} introduced a novel metric targeting the quantification of overlap of clinical entities between ground truth and generated reports in chest X-ray report generation. They use the RadGraph model \cite{radgraph}, a language model trained on a limited subset of reports from the MIMIC-CXR dataset \cite{mimic}. The MIMIC-CXR dataset consists of chest X-ray images with corresponding reports, and the RadGraph dataset includes medical entities from chest X-ray reports annotated by radiologists.
The approach by Yu et al. is similar to the BLEU score, exclusively considering the exact matches among the primary entities in generated and ground truth reports, overlooking the semantic similarity of these entities. Furthermore, the generalizability of this approach to other medical applications is constrained by the RadGraph model's specialization in extracting only chest X-ray related entities. Nonetheless, while the RadGraph model acknowledges negations in the texts, they are treated merely as labels to the entities, and the details of entity descriptions are not factored into the evaluation process.

In a recent study, Patricoski et al. \cite{patricoski2022evaluation-Bert} conducted an evaluation of seven BERT models to assess semantic similarity in clinical trial texts. Notably, the pre-trained BERT model known as SciBERT \cite{beltagy-etal-2019-scibert} demonstrated better performance compared to the other BERT models, even outperforming the standard BERT model, which secured the second position in this evaluation. This study underlines the promising potential of BERT models in semantic similarity evaluation. However, it has a drawback associated with using BERT models without preprocessing.
BERT models operate at a token-by-token level, evaluating semantic similarity by comparing all tokens with each other, a computationally intensive process that gives relatively low scores. Despite this computational challenge, it is important to consider the significant potential in SciBERT, particularly due to its huge clinical dictionary. This finding underscores the need for careful consideration of preprocessing strategies to maximize the effectiveness of BERT models in semantic similarity evaluations.

Notably, the absence of a comprehensive, general semantic similarity evaluation metric for medical content persists. Consequently, we introduce a novel metric for Medical Corpus Similarity Evaluation (MCSE) to comprehensively address and resolve these challenges.
\section{\uppercase{Methodology}}
\label{sec:method}

We developed a novel metric for Medical Corpus Similarity Evaluation (MCSE), by exclusively extracting key medical entities and employing a pretrained BERT model to assess the semantic similarity of these entities within chest X-ray reports. This targeted approach allows BERT to concentrate solely on important information and reduces the computational load during comparison. Importantly, our methodology goes beyond extracting main entities, we also consider the negations and detailed descriptions associated with the primary medical entities in chest X-ray reports. Our MCSE metric consists of two essential steps: 
\begin{enumerate}
    \item Clinical Entity Extraction.
    \item Domain Similarity Evaluation.
\end{enumerate}

\subsection{Clinical Entity Extraction}
The most important part of comprehending semantic similarity evaluation in text relies on identifying the key elements, often referred to as clinical entities, within medical texts. These entities typically fall into categories related to anatomical body parts, symptoms, laboratory equipment, and diagnoses. Each category is typically signaled by certain words within a sentence. However, there are additional words that precede or follow these main entities, offering descriptions.

To address these complexities, we employ the Scispacy model \cite{neumann-etal-2019-scispacy} for extracting primary clinical entities from medical text using the embedded clinical dictionary in this model (BC5CDR: a corpus comprising 1500 PubMed articles with 4409 annotated chemicals, 5818 diseases, and 3116 chemical-disease interactions \cite{bc5cdr}). Subsequently, we automatically process the entire text to identify associated negations and adjectives related to these key entities. These elements are then integrated to provide a comprehensive representation of the considered text. In the context of this research, the category of laboratory equipment is deliberately excluded, aligning with the specific focus of our application. Table \ref{tab:example_entity} presents an example of medical text and the extracted entities using our method and the Scispacy method without any cleaning process.
While we employ the Scispacy model for initial entity extraction, it is evident that this model alone may not suffice. An additional automated post-processing step is needed to refine and integrate related entities. 
The post-processing steps involve eliminating a single adjective or non-medical entities, excluding entities categorized as lab equipment, identifying and adding the relevant adjective to the remaining medical entities, including the existing negation into these primary entities, and screening out any reported diagnostic procedures terms.
These processes are essential to ensure that the final output is presented as a cohesive set of primary medical entities, ready for practical use.

\begin{table*}[]
    \centering
    \caption{In the right column there is an example of medical text. In the left column, there are clinical entities extracted using the Scispacy model without any cleaning process, and In the middle column, there are clinical entities extracted using our method.}
    \begin{tabular}{|p{3in}|p{1in}|p{1.5in}|}
    \hline
    Medical Text & Extracted Entities using our method & Extracted Entities using Scispacy \cite{neumann-etal-2019-scispacy}\\
    \hline
         1. Interval clearance of left basilar consolidation. 2. Patchy right basilar opacities, which could be seen with minor atelectasis, but given the context clinical correlation is suggested regarding any possibility for recurrent or new aspiration pneumonitis at the right lung base. 3. Increased new interstitial abnormality, suggesting recurrence of fluid overload or mild-to-moderate pulmonary edema; aspiration could also be considered. Inflammation associated with atypical infectious process is probably less likely given the waxing and waning presentation.& fluid overload, inflammation, aspiration pneumonitis, minor atelectasis, mild to moderate pulmonary edema, left basilar consolidation, patchy right basilar opacities, interstitial abnormality & Interval, clearance, left basilar, consolidation, Patchy, right basilar, opacities, minor, atelectasis, clinical, recurrent, aspiration, pneumonitis, right lung base, Increased, interstitial abnormality, recurrence, fluid, overload, mild-to-moderate pulmonary edema, aspiration, Inflammation, associated with, atypical, infectious process, waxing, waning, presentation\\
        \hline
    \end{tabular}
    \label{tab:example_entity}
\end{table*}

\subsection{Domain Similarity Evaluation}

Having successfully extracted and shifted our focus to the primary entities within the medical corpus, the next step involves assessing their semantic similarities by assigning corresponding scores.

After processing entity extraction, we calculate a similarity score for the sequences of entities. Let $T = (t_1, \dots, t_N)$ represent the reference text entities and $\hat{T} = (\hat{t}_1, \dots, \hat{t}_M)$ represent the generated text or candidate text entities. Initially, we identify the exact same medical entities in both sequences and determine the total count ($|C^{(i)}|$). For the remaining entities, we construct a similarity matrix, where each element represents the similarity score between entities, as illustrated in table \ref{tab:sm_example}.

\begin{gather}
    S_i = \frac{\max y_{i,j}}{\max y_{i,j} + \overline{y_{i,j}}}  \mkern12mu i = (0, 1, \dots, M) \mkern3mu j = (0, 1, \dots, N) 
    \label{eq:similarity1}
    \\
    y_{i,j} = Similarity(r_{i}, \hat{r}_{j})
    \label{eq:similarity2}
\end{gather}
\begin{equation}
    \begin{cases}
   C^{(i)} =  t_i, & \text{if } t_i = \hat{t}_j\\
   r_{i} = t_i \And \hat{r}_{j} = \hat{t}_j & \text{if } t_i \neq \hat{t}_j
\end{cases}
 \label{eq:nomatch}
\end{equation}
Where $M$ is the number of total candidate entities, $r_{i}$ and $\hat{r}_{j}$ are the sequence between no matched entities as in equation \eqref{eq:nomatch},  and  $S_i$ is a normalized similarity score between $r_{i}$ and $\hat{r}_{j}$.
The similarity score $Similarity(r_{i}, \hat{r}_{j})$ in equation \eqref{eq:similarity2} is derived from spaCy \cite{Honnibal_spaCy_Industrial-strength_Natural_2020}, a BERT model trained on word2vec, to evaluate similarities using domain cosine similarity.

To evaluate the similarity of candidate entities with the reference entities, we compute the maximum score for each column and normalized it with the column's average ($S_i$). We then sum these scores for each column, adding them to $|C^{(i)}|$. To obtain the final similarity score between the two corpora, we divide this sum by the total number of candidate entities. This process is explained in Equation \eqref{eq:mcse}.

\begin{equation}
    MCSE := \frac{|C^{(i)}| + \sum_{i=1}^M S_i}{M}
    \label{eq:mcse}
\end{equation}

Where $|C^{(i)}|$ is the number of exactly matched entities between the two corpora of $T$ and $\hat{T}$.

For instance, Table \ref{tab:sm_example} provides an example of the probable similarity score that two sets of entities can receive. These entities have been extracted using our medical entity extraction procedure.

In the table, the two corpora received a score of 0.55 according to our MCSE metric. However, the calculated BLEU score for them is approximately zero.
Upon analyzing the two medical texts, it becomes evident that although the candidate text does refer to the same side of the chest as in the reference text and that both texts indicate the presence of pulmonary edema and pulmonary masses, their overall similarity is relatively limited. The score of 0.55 carries a more meaningful value in this context compared to the nearly zero score generated by BLEU.

\begin{table*}[h]
    \caption{An example of a medical similarity score between entities. Each score is calculated from equation \eqref{eq:similarity2}, with the final row $S_i$ being computed using equation \eqref{eq:similarity1}. The scores highlighted in blue indicate the maximum value within each respective column.}
    \begin{tabular}{|p{0.6in}|p{1.7in}|ccc|}
    \hline
        \multicolumn{5}{|p{6in}|}{\textbf{Reference:} 1. Interval clearance of left basilar consolidation. 2. Patchy right basilar opacities, which could be seen with minor atelectasis, but given the context clinical correlation is suggested regarding any possibility for recurrent or new aspiration pneumonitis at the right lung base. 3. Increased new interstitial abnormality, suggesting recurrence of fluid overload or mild-to-moderate pulmonary edema; aspiration could also be considered. Inflammation associated with atypical infectious process is probably less likely given the waxing and waning presentation.}  \\
        \hline
         \multicolumn{5}{|p{6in}|}{\textbf{Candidate:} Stable multiple bilateral pulmonary masses and right middle lobe collapse due to hilar adenopathy.} \\
         \hline
         \multicolumn{2}{|c|}{}&\multicolumn{3}{|c|}{Candidate Medical Entities}\\
         \cline{3-5}
          \multicolumn{2}{|c|}{} &pulmonary masses&right middle lobe&hilar adenopathy\\
          \hline
          &fluid overload&0.61&0.49&0.45\\
          \cline{2-5}
          &inflammation&0.64&0.48&0.55\\
          \cline{2-5}
          &aspiration pneumonitis&0.65&0.39&0.50\\
          \cline{2-5}
          &minor atelectasis&0.62&0.47&0.53\\
          \cline{2-5}
          Reference Medical Entities&mild to moderate pulmonary edema&\textbf{\textcolor{blue}{0.78}}&0.31&0.51\\
          \cline{2-5}
          &left basilar consolidation&0.52&\textbf{\textcolor{blue}{0.66}}&0.32\\
          \cline{2-5}
          &patchy right basilar opacities&0.64&\textbf{\textcolor{blue}{0.66}}&0.49\\
          \cline{2-5}
          &interstitial abnormality&0.69&0.63&\textbf{\textcolor{blue}{0.59}}\\
          \hline
           \multicolumn{2}{|c|}{$S_i$}&0.548&0.563&0.545\\
           \hline
    \end{tabular}
    \label{tab:sm_example}
\end{table*}
\section{\uppercase{Validation}}
\label{sec:validation}

While the underlying logic of this metric is reasonable, it is imperative that we validate the results robustly. Given the use of chest X-ray reports for this particular application, we have conducted an extensive search within existing datasets to identify an appropriate validation method.
After a comprehensive review of various datasets, we concluded that it would be more effective to conduct separate validations for the different steps of the proposed metric.

\subsection{Clinical Entity Extraction Process}
In order to rigorously validate our clinical entity extraction process, we employ the RadGraph dataset \cite{radgraph}. This dataset is a valuable resource in which radiologists thoroughly annotated the primary clinical entities in chest X-ray reports as either "definitely present" within the report or "definitely absent". Importantly, in cases where a negation is associated with a particular entity, it is annotated as "definitely absent."

To achieve our validation objectives, we executed our entity extraction process on the reports within this dataset. Subsequently, we compare the number of similar entities extracted through our method with the annotations provided by radiologists, particularly focusing on the two categories of "definitely present" and "definitely absent". This systematic comparison allows us to assess the accuracy and effectiveness of our clinical entity extraction methodology in the context of chest X-ray reports, aligning with radiological standards.
Throughout the validation process, covering all reports in our study, our method consistently achieves a high level of accuracy. On average, it accurately recognizes 75\% of entities marked as "definitely present" and successfully identifies 76\% of entities labeled as "definitely absent".
In our entity extraction process, we deliberately omit anatomical entities like "chest" or "lung," as they are redundant to the chest X-ray application and do not contribute significantly to the process. This selective exclusion is one of the factors contributing to the approximately 75\% accuracy in our results. Nevertheless, these results affirm the reliability and consistency of our methodology. 

\subsection{Domain Similarity Score}

In contrast to the initial phase of clinical entity extraction, validating the domain similarity score is more challenging. The scoring system itself is more controversial and subject to debate, and creating an automated validation method, free from reliance on radiologists, necessitates a creative and innovative approach.
Nevertheless, through the available tools and databases, we establish a dedicated system for the validation of this scoring method for the application of chest X-rays.

In the chest X-ray application, the MIMIC-CXR dataset \cite{mimic}, is one of the biggest available databases for chest X-ray images and their corresponding reports. Notably, this dataset provides us with Chexpert labels (Medical Observation), including Atelectasis, Cardiomegaly, Consolidation, Edema, Enlarged Cardiomediastinum, Fracture, Lung Lesion, Lung Opacity, No Finding, Pleural Effusion, Pleural Other, Pneumonia, Pneumothorax, and Support Devices labels \cite{irvin2019chexpert}. The values of each label are 1 (definitely present), 0 (definitely absent), -1 (ambiguous), or it carries no value at all. Table \ref{tab:chexpert_mimic} presents a sample of Chexpert labels extracted from chest X-ray reports of five patients from the MIMIC-CXR database. The reports corresponding to these subjects are presented in Table \ref{tab:chexpert_mimic_report}.

\begin{table*}[]
    \centering
    \caption{A sample table featuring Chexpert labels (1. Atelectasis, 2. Cardiomegaly, 3. Consolidation, 4. Edema, 5. Enlarged Cardiomediastinum, 6. Fracture, 7. Lung Lescion, 8. Lung Opacity, 9. No Finding, 10. Pleural Effusion, 11. Pleural Other, 12. Pneumonia, 13. Pneumothorax, 14. Support Devices) extracted from chest X-ray reports of five patients (Subject \#\#) from the MIMIC-CXR database \cite{mimic}.}
    \begin{adjustbox}{width=1\textwidth}
        \begin{tabular}{|c|c|c|c|c|c|c|c|c|c|c|c|c|c|c|}
        \hline
             Subject \#\#&Atelectasis&Cardiomegaly&Consolidation&Edema&Enlarged Cardiomediastinum&Fracture&Lung Lescion&Lung Opacity&No Finding&Pleural Effusion&Pleural Other&Pneumonia&Pneumothorax&Support Devices \\
             \hline
             01&&&&&&&&0&&1&&1&-1&\\
             \hline
            02&&&&&&&1&&&&&1&&\\
             \hline
             03&&&&&&&&&1&&&0&&\\
             \hline
             04&&1&0&&&&&-1&&&&&0&1\\
             \hline
             05&1&&&&&&&&&1&&&&\\
             \hline
        \end{tabular}
    \end{adjustbox}
    \label{tab:chexpert_mimic}
\end{table*}
\begin{table}[]
    \centering
    \caption{Reports corresponding to the subjects listed in Table \ref{tab:chexpert_mimic} from the MIMIC-CXR dataset \cite{mimic}.}   
    \scriptsize
    \begin{tabular}{|p{0.35in}|p{2.2in}|}
    \hline
        Subject\_\#\# & Report \\
        \hline
        01 &  Lung volumes remain low.  There are innumerable bilateral scattered small pulmonary nodules which are better demonstrated on recent CT.  Mild pulmonary vascular congestion is stable.  The cardio mediastinal silhouette and hilar contours are unchanged.  Small pleural effusion in the right middle fissure is new.  There is no new focal opacity to suggest pneumonia.  There is no pneumothorax.\\
        \hline
        02 & A triangular opacity in the right lung apex is new from prior examination.  There is also fullness of the right hilum which is new. The remainder of the lungs are clear.  Blunting of bilateral costophrenic angles, right greater than left, may be secondary to small effusions.  The heart size is top normal.\\
        \hline
        03 &  Mild to moderate enlargement of the cardiac silhouette is unchanged.  The aorta is calcified and diffusely tortuous.  The mediastinal and hilar contours are otherwise similar in appearance.  There is minimal upper zone vascular redistribution without overt pulmonary edema.  No focal consolidation, pleural effusion or pneumothorax is present.  The osseous structures are diffusely demineralized.\\
        \hline
        04 & The endotracheal tube tip is 6 cm above the carina.  Nasogastric tube tip is beyond the GE junction and off the edge of the film.  A left central line is present in the tip is in the mid SVC.  A pacemaker is noted on the right in the lead projects over the right ventricle.  There is probable scarring in both lung apices.  There are no new areas of consolidation.  There is upper zone redistribution and cardiomegaly suggesting pulmonary venous hypertension. There is no pneumothorax.\\
        \hline
        05 & A moderate left pleural effusion is new. Associated left basilar opacity likely reflect compressive atelectasis.  There is no pneumothorax.  There are no new abnormal cardiac or mediastinal contour. Median sternotomy wires and mediastinal clips are in expected positions.\\
        \hline
    \end{tabular}
    \label{tab:chexpert_mimic_report}
\end{table}
Our approach involves two distinct strategies. Firstly, we seek to identify reports sharing the same sequence of labels and values. 
For instance, we search for reports from subjects with Chexpert label sequences similar to that of Subject\_01 in Table \ref{tab:chexpert_mimic}.
For these reports with matching label sequences, we proceed to similarity scores computation for each pair of reports. 
Simultaneously, we identify reports featuring only one or two labels and with a value of "definitely present" for these labels resembling Subject\_02 in Table \ref{tab:chexpert_mimic} and assess the similarity of these reports with the reports with different label sequences. 
As an example, we calculate the similarity between the reports of Subject\_02 and Subject\_05 from Table \ref{tab:chexpert_mimic}, given their entirely distinct label sequences.
This two-fold method allows us to analyze the semantic similarity scores for both similar and contrasting reports in terms of their labels.

\begin{figure}
    \centering
    \includegraphics[trim={1cm 0cm 1cm 0cm}, width=0.48\textwidth, clip]{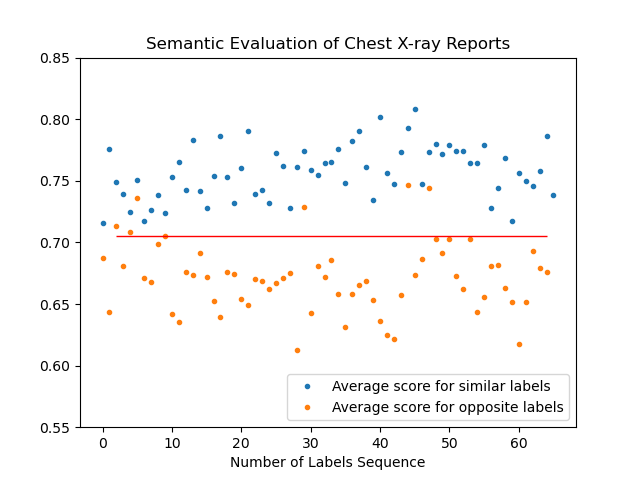}
    \caption{Semantic Evaluation of Chest X-ray reports. Each blue dot represents the mean score of semantic evaluation for reports with similar label sequences, while each orange dot signifies the mean score of semantic evaluation for reports with opposing labels. The red horizontal line represents the classification boundary.}
    \label{fig:validation}
\end{figure}

Figure \ref{fig:validation} presents the results of the two-fold validation for our scoring method. Within the figure, blue dots represent the average scores for semantic evaluation of reports with similar label sequences, while orange dots show the mean scores for reports with contrasting labels. The red horizontal line within the figure serves as the dividing line distinguishing between similar and opposite evaluations.
Upon reviewing these results, it becomes evident that a distinct boundary exists between reports sharing the same clinical diagnoses and those with entirely dissimilar diagnoses. Notably, there are no blue dots below a 70\% similarity threshold, whereas six orange dots have scores above 70\% across 70 label sequences, which is certainly not very high.
Nevertheless, despite this differentiation between similar and opposite evaluations, some level of similarity, exceeding 50\%, persists within the opposing category. This can be attributed to the implemented cosine similarity within the medical domain, which introduces a certain bias towards tokens in the same medical domain. Unfortunately, this bias cannot be entirely eliminated, as it plays a substantial role in the evaluation process. However, a clear boundary remains between similar and contrasting reports. 
\section{\uppercase{Results and Discussion}}
\label{sec:result}

In our original application of chest X-ray report generation, we incorporate our metric to assess the outputs of various models. We compare our results with the BLEU scores evaluated by these models, specifically, the CXR-RePaiR \cite{cxr-repair-endo21a} and R2Gen \cite{chen2022generating} models, both being state-of-the-art models for generating chest X-ray reports. Our evaluation focuses on measuring the semantic similarity between the generated reports and the ground truth. Table \ref{tab:sota_bleu} presents the BLEU scores obtained from these models and our metric's semantic evaluation.
As anticipated, the BLEU scores are relatively low, signifying a substantial dissimilarity between the generated results and the ground truth for both the CXR-RePaiR and R2Gen models despite being regarded as state-of-the-art models for chest X-ray report generation. These models still employ the BLEU metric for evaluation, primarily due to the scarcity of more suitable metrics and the need for a standardized evaluation process for comparative purposes.
Conversely, our metric produces more promising results for both of these models. While our metric's scores align with the BLEU scores, indicating higher scores for both BLEU and our MCSE metric in the case of the R2Gen model compared to the CXR-RePaiR, our metric provides a deeper evaluation. It suggests a degree of similarity to the ground truth rather than outright dissimilarity in BLEU, thus making the generated reports more reliable and trustworthy, which is a crucial advancement in the field.

\begin{table}[h]
    \centering
    \caption{The result of BLEU score of 2-gram for state-of-the-art models and the result of our novel metric on these models outcomes.}
    \begin{tabular}{|p{1.5in}|c|c|}
    \hline
         \hspace{1.2cm} Models & BLEU & Our MCSE \\
         \hline
         R2Gen \cite{chen2022generating} & 0.212 & 0.71\\
         CXR-RePair \cite{cxr-repair-endo21a} & 0.069 & 0.64\\
         \hline
    \end{tabular}
    \label{tab:sota_bleu}
\end{table}

Table \ref{tab:example_metric} provides an example of medical text generated and evaluated using both a BLEU score and our MCSE metric. It's evident that, according to the BLEU score, these two texts appear vastly different, even though they share the same primary medical entities. However, when we delve into the context, we can notice that "moderately severe" serves as a description for the main entity, "pulmonary edema", in the generated text. Similarly, in the second part of the text, the main medical entity is "pleural effusions", and terms like "likely" and "no large" are used to describe this entity, which may not be identical but share semantic similarities. This subtle context evaluation is precisely what our metric considers, yielding a similarity score of 0.64 for these texts, which we argue is a more accurate reflection compared to the BLEU score.

\begin{table}[]
    \centering
    \caption{A comparative example of using the BLEU score and our adapted metric with medical reference and generated text.}
    \begin{center}
     \begin{tabular}{|p{1.7in}|c|c|}
    \hline
    &BLEU&MCSE\\
    \hline
        \textbf{Reference Sentence:} "Pulmonary edema, cardiomegaly, likely pleural effusions."
        
        \noindent\textbf{Generated Sentence: }"Moderately severe bilateral pulmonary edema with no large pleural effusion." & 0.047 & \textbf{0.64}   \\
         \hline
    \end{tabular}   
    \end{center}
    
    \label{tab:example_metric}
\end{table}

Lastly, the significant benefit of employing this metric lies in its capacity for comparative analysis alongside other evaluation measures. For instance, when examining the outcomes of the BLEU score, with its word-by-word analysis, situations may arise where the results are totally inaccurate, casting doubt on their reliability, despite the models performing well overall. Integrating the results of our novel MCSE metric into the evaluation process allows us to semantically analyze and ascertain the dependability of the models' textual outputs within the context of medical content.
\section{\uppercase{Conclusion}}
\label{sec:conclusion}

In our research, we tackle the challenge of semantic similarity scoring in medical corpora, driven by the inadequacy of existing metrics that, while suitable for machine translation evaluation, fall short in the field of medical semantic assessment. Our innovative metric draws inspiration from how humans comprehend text, centering on the extraction of key terms and their relational context. It introduces a novel approach for extracting clinical entities from medical text, considering not only the entities themselves but also the associated descriptions and negations. Additionally, we created a  new method for scoring the semantic relationships between these entities by using the domain cosine similarity.
The validation process allowed us to analyze and validate each of these steps individually, unraveling a clear distinction between reports sharing the same diagnosis and those diverging in this regard.

For our research, we focused on the application of chest X-rays, a critical domain where a robust semantic evaluation metric is highly valuable. We applied our metric to some of the latest state-of-the-art models, and the results harmonized with other evaluation metrics, affirming their reliability.

While our validation process and implementation yielded successful outcomes, we encountered the challenge of an inherent bias in domain cosine similarity. This challenge has illuminated a promising direction for our future research, as we explore ways to mitigate this bias and advance the field of medical semantic evaluation.
\paragraph*{\textbf{Material, codes, and Acknowledgement:}} Results can be reproduced using the code available in the GitHub repository \url{https://github.com/sayeh1994/Medical-Corpus-Semantic-Similarity-Evaluation.git}. All the computations presented in this paper were performed using the \cite{gricad} infrastructure (\url{https://gricad.univ-grenoble-alpes.fr}), which is supported by Grenoble research communities.

\bibliographystyle{unsrtnat}
\bibliography{example}

\begin{thebibliography}{18}
\providecommand{\natexlab}[1]{#1}
\providecommand{\url}[1]{\texttt{#1}}
\expandafter\ifx\csname urlstyle\endcsname\relax
  \providecommand{\doi}[1]{doi: #1}\else
  \providecommand{\doi}{doi: \begingroup \urlstyle{rm}\Url}\fi

\bibitem[Alam et~al.(2020)Alam, Afzal, and Malik]{intro1}
Fakhare Alam, Muhammad Afzal, and Khalid~Mahmood Malik.
\newblock Comparative analysis of semantic similarity techniques for medical text.
\newblock In \emph{2020 International Conference on Information Networking (ICOIN)}, pages 106--109, 2020.
\newblock \doi{10.1109/ICOIN48656.2020.9016574}.

\bibitem[Endo et~al.(2021)Endo, Krishnan, Krishna, Ng, and Rajpurkar]{cxr-repair-endo21a}
Mark Endo, Rayan Krishnan, Viswesh Krishna, Andrew~Y. Ng, and Pranav Rajpurkar.
\newblock Retrieval-based chest x-ray report generation using a pre-trained contrastive language-image model.
\newblock In \emph{Proceedings of Machine Learning for Health}, volume 158 of \emph{Proceedings of Machine Learning Research}, pages 209--219, 2021.

\bibitem[Miura et~al.(2021)Miura, Zhang, Tsai, Langlotz, and Jurafsky]{miura-etal-2021-improving-m2trans}
Yasuhide Miura, Yuhao Zhang, Emily Tsai, Curtis Langlotz, and Dan Jurafsky.
\newblock Improving factual completeness and consistency of image-to-text radiology report generation.
\newblock In Kristina Toutanova, Anna Rumshisky, Luke Zettlemoyer, Dilek Hakkani-Tur, Iz~Beltagy, Steven Bethard, Ryan Cotterell, Tanmoy Chakraborty, and Yichao Zhou, editors, \emph{Proceedings of the 2021 Conference of the North American Chapter of the Association for Computational Linguistics: Human Language Technologies}, pages 5288--5304, Online, June 2021. Association for Computational Linguistics.
\newblock \doi{10.18653/v1/2021.naacl-main.416}.
\newblock URL \url{https://aclanthology.org/2021.naacl-main.416}.

\bibitem[Yu et~al.(2022)Yu, Endo, Krishnan, Pan, Tsai, Reis, Fonseca, Ho~Lee, Abad, Ng, Langlotz, Venugopal, and Rajpurkar]{yuEvaluatingProgressAutomatic2022}
Feiyang Yu, Mark Endo, Rayan Krishnan, Ian Pan, Andy Tsai, Eduardo~Pontes Reis, Eduardo Kaiser Ururahy~Nunes Fonseca, Henrique~Min Ho~Lee, Zahra Shakeri~Hossein Abad, Andrew~Y. Ng, Curtis~P. Langlotz, Vasantha~Kumar Venugopal, and Pranav Rajpurkar.
\newblock Evaluating {Progress} in {Automatic} {Chest} {X}-{Ray} {Radiology} {Report} {Generation}.
\newblock preprint, Radiology and Imaging, August 2022.
\newblock URL \url{http://medrxiv.org/lookup/doi/10.1101/2022.08.30.22279318}.

\bibitem[Chen et~al.(2020)Chen, Song, Chang, and Wan]{chen2022generating}
Zhihong Chen, Yan Song, Tsung-Hui Chang, and Xiang Wan.
\newblock Generating radiology reports via memory-driven transformer.
\newblock In \emph{Proceedings of the 2020 Conference on Empirical Methods in Natural Language Processing (EMNLP)}, pages 1439--1449, Online, November 2020. Association for Computational Linguistics.
\newblock \doi{10.18653/v1/2020.emnlp-main.112}.
\newblock URL \url{https://aclanthology.org/2020.emnlp-main.112}.

\bibitem[Papineni et~al.(2002)Papineni, Roukos, Ward, and Zhu]{papineni-etal-2002-bleu}
Kishore Papineni, Salim Roukos, Todd Ward, and Wei-Jing Zhu.
\newblock {B}leu: a method for automatic evaluation of machine translation.
\newblock In \emph{Proceedings of the 40th Annual Meeting of the Association for Computational Linguistics}, pages 311--318, Philadelphia, Pennsylvania, USA, July 2002. Association for Computational Linguistics.
\newblock \doi{10.3115/1073083.1073135}.
\newblock URL \url{https://aclanthology.org/P02-1040}.

\bibitem[Banerjee and Lavie(2005)]{banerjee-lavie-2005-meteor}
Satanjeev Banerjee and Alon Lavie.
\newblock {METEOR}: An automatic metric for {MT} evaluation with improved correlation with human judgments.
\newblock In \emph{Proceedings of the {ACL} Workshop on Intrinsic and Extrinsic Evaluation Measures for Machine Translation and/or Summarization}, pages 65--72, Ann Arbor, Michigan, June 2005. Association for Computational Linguistics.
\newblock URL \url{https://aclanthology.org/W05-0909}.

\bibitem[Lin(2004)]{lin-2004-rouge}
Chin-Yew Lin.
\newblock {ROUGE}: A package for automatic evaluation of summaries.
\newblock In \emph{Text Summarization Branches Out}, pages 74--81, Barcelona, Spain, July 2004. Association for Computational Linguistics.
\newblock URL \url{https://aclanthology.org/W04-1013}.

\bibitem[Smit et~al.(2020)Smit, Jain, Rajpurkar, Pareek, Ng, and Lungren]{Smit2020CheXbertCA}
Akshay Smit, Saahil Jain, Pranav Rajpurkar, Anuj Pareek, A.~Ng, and Matthew~P. Lungren.
\newblock Chexbert: Combining automatic labelers and expert annotations for accurate radiology report labeling using bert.
\newblock In \emph{Conference on Empirical Methods in Natural Language Processing}, 2020.
\newblock URL \url{https://api.semanticscholar.org/CorpusID:215827807}.

\bibitem[Irvin et~al.(2019)Irvin, Rajpurkar, Ko, Yu, Ciurea-Ilcus, Chute, Marklund, Haghgoo, Ball, Shpanskaya, Seekins, Mong, Halabi, Sandberg, Jones, Larson, Langlotz, Patel, Lungren, and Ng]{irvin2019chexpert}
Jeremy Irvin, Pranav Rajpurkar, Michael Ko, Yifan Yu, Silviana Ciurea-Ilcus, Chris Chute, Henrik Marklund, Behzad Haghgoo, Robyn Ball, Katie Shpanskaya, Jayne Seekins, David~A. Mong, Safwan~S. Halabi, Jesse~K. Sandberg, Ricky Jones, David~B. Larson, Curtis~P. Langlotz, Bhavik~N. Patel, Matthew~P. Lungren, and Andrew~Y. Ng.
\newblock Chexpert: A large chest radiograph dataset with uncertainty labels and expert comparison.
\newblock \emph{Proceedings of the AAAI Conference on Artificial Intelligence}, 33\penalty0 (01):\penalty0 590--597, Jul. 2019.
\newblock \doi{10.1609/aaai.v33i01.3301590}.
\newblock URL \url{https://ojs.aaai.org/index.php/AAAI/article/view/3834}.

\bibitem[Jain et~al.(2021)Jain, Agrawal, Saporta, Truong, Duong, Bui, Chambon, Zhang, Lungren, Ng, Langlotz, and Rajpurkar]{radgraph}
Saahil Jain, Ashwin Agrawal, Adriel Saporta, Steven Truong, Du~Nguyen Duong, Tan Bui, Pierre Chambon, Yuhao Zhang, Matthew~P. Lungren, Andrew~Y. Ng, Curtis Langlotz, and Pranav Rajpurkar.
\newblock Radgraph: Extracting clinical entities and relations from radiology reports.
\newblock In \emph{Thirty-fifth Conference on Neural Information Processing Systems Datasets and Benchmarks Track (Round 1)}, 2021.
\newblock URL \url{https://openreview.net/forum?id=pMWtc5NKd7V}.

\bibitem[Johnson et~al.(2019)Johnson, Pollard, Berkowitz, Greenbaum, Lungren, Deng, Mark, and Horng]{mimic}
Alistair E.~W. Johnson, Tom~J. Pollard, Seth~J. Berkowitz, Nathaniel~R. Greenbaum, Matthew~P. Lungren, Chih-ying Deng, Roger~G. Mark, and Steven Horng.
\newblock Mimic-cxr, a de-identified publicly available database of chest radiographs with free-text reports.
\newblock \emph{Scientific Data}, 6\penalty0 (1):\penalty0 317, Dec 2019.
\newblock ISSN 2052-4463.
\newblock \doi{10.1038/s41597-019-0322-0}.
\newblock URL \url{https://doi.org/10.1038/s41597-019-0322-0}.

\bibitem[Patricoski et~al.(2022)Patricoski, Kreimeyer, Balan, Hardart, Tao, Anagnostou, Botsis, Investigators, et~al.]{patricoski2022evaluation-Bert}
Jessica Patricoski, Kory Kreimeyer, Archana Balan, Kent Hardart, Jessica Tao, Valsamo Anagnostou, Taxiarchis Botsis, Johns Hopkins Molecular Tumor~Board Investigators, et~al.
\newblock An evaluation of pretrained bert models for comparing semantic similarity across unstructured clinical trial texts.
\newblock \emph{Stud Health Technol Inform}, 289:\penalty0 18--21, 2022.

\bibitem[Beltagy et~al.(2019)Beltagy, Lo, and Cohan]{beltagy-etal-2019-scibert}
Iz~Beltagy, Kyle Lo, and Arman Cohan.
\newblock {S}ci{BERT}: A pretrained language model for scientific text.
\newblock In Kentaro Inui, Jing Jiang, Vincent Ng, and Xiaojun Wan, editors, \emph{Proceedings of the 2019 Conference on Empirical Methods in Natural Language Processing and the 9th International Joint Conference on Natural Language Processing (EMNLP-IJCNLP)}, pages 3615--3620, Hong Kong, China, November 2019. Association for Computational Linguistics.
\newblock \doi{10.18653/v1/D19-1371}.
\newblock URL \url{https://aclanthology.org/D19-1371}.

\bibitem[Neumann et~al.(2019)Neumann, King, Beltagy, and Ammar]{neumann-etal-2019-scispacy}
Mark Neumann, Daniel King, Iz~Beltagy, and Waleed Ammar.
\newblock {S}cispa{C}y: {F}ast and {R}obust {M}odels for {B}iomedical {N}atural {L}anguage {P}rocessing.
\newblock In \emph{Proceedings of the 18th BioNLP Workshop and Shared Task}, pages 319--327, Florence, Italy, August 2019. Association for Computational Linguistics.
\newblock \doi{10.18653/v1/W19-5034}.
\newblock URL \url{https://www.aclweb.org/anthology/W19-5034}.

\bibitem[Li et~al.(2016)Li, Sun, Johnson, Sciaky, Wei, Leaman, Davis, Mattingly, Wiegers, and Lu]{bc5cdr}
Jiao Li, Yueping Sun, Robin~J. Johnson, Daniela Sciaky, Chih-Hsuan Wei, Robert Leaman, Allan~Peter Davis, Carolyn~J. Mattingly, Thomas~C. Wiegers, and Zhiyong Lu.
\newblock {BioCreative V CDR task corpus: a resource for chemical disease relation extraction}.
\newblock \emph{Database}, 2016:\penalty0 baw068, 05 2016.
\newblock ISSN 1758-0463.
\newblock \doi{10.1093/database/baw068}.
\newblock URL \url{https://doi.org/10.1093/database/baw068}.

\bibitem[Honnibal et~al.(2020)Honnibal, Montani, Van~Landeghem, and Boyd]{Honnibal_spaCy_Industrial-strength_Natural_2020}
Matthew Honnibal, Ines Montani, Sofie Van~Landeghem, and Adriane Boyd.
\newblock {spaCy: Industrial-strength Natural Language Processing in Python}.
\newblock 2020.
\newblock \doi{10.5281/zenodo.1212303}.

\bibitem[Gricad()]{gricad}
Gricad.
\newblock infrastructure supported by grenoble research communities.
\newblock URL \url{https://gricad.univ-grenoble-alpes.fr}.

\end{thebibliography}
\end{document}